\documentclass[letterpaper, 10 pt, conference]{ieeeconf}  
\IEEEoverridecommandlockouts                              

\usepackage{graphics} 
\usepackage{epsfig} 

\usepackage{stfloats}
\usepackage{times} 
\usepackage{amsmath} 
\usepackage{amssymb}  
\usepackage[hang]{subfigure} 
\usepackage{cite}
\usepackage{siunitx}
\usepackage{soul}
\usepackage{color}
\usepackage{xcolor}
\usepackage{mathrsfs}
\usepackage[ruled,linesnumbered]{algorithm2e}
\usepackage[font=scriptsize]{caption}

\soulregister\ref7  
\soulregister\cite7 
\soulregister\eqref7
\soulregister\prettyref7

\usepackage{multirow}
\usepackage{algpseudocode}
\usepackage{makecell}
\usepackage{booktabs}
\usepackage{threeparttable}
\usepackage{hyperref}
\usepackage{endnotes}

\bstctlcite{IEEEexample:BSTcontrol}

\usepackage{prettyref}
\newrefformat{fig}{Fig.~\ref{#1}}
\newrefformat{sec}{Sec.~\ref{#1}}
\newrefformat{tab}{Tab.~\ref{#1}}
\newrefformat{algo}{Algo.~\ref{#1}}
\newrefformat{eq}{(\ref{#1})}

\usepackage{color}

\title{\LARGE \bf
\textcolor{magenta}{Trinity}: A Modular Humanoid Robot AI System}

\author{Jingkai Sun$^{1,2,\ast}$, Qiang Zhang$^{1,2,\ast,\clubsuit}$, Gang Han$^{1,\ast}$, Wen Zhao$^{1,\ast}$, \\
Zhe Yong$^{1}$, Yan He$^{1}$, Jiaxu Wang$^{2}$, Jiahang Cao$^{2}$, Yijie Guo$^{1,\clubsuit}$, Renjing Xu$^{2,\dagger}$}

\begin{document}
\maketitle

\footnotetext[1]{The authors are with Beijing Innovation Center of Humanoid Robotics Co. Ltd. {\tt\small Jony.Zhang@xhumanoid.com}}
\footnotetext[2]{The authors are with The Hong Kong University of Science and Technology (Guangzhou), China. {$^{\ast}$ Equal contribution; $^{\dagger}$ Corresponding author; $^{\clubsuit}$ Project leader.} {\tt\small  \ qzhang749@connect.hkust-gz.edu.cn, renjingxu@hkust-gz.edu.cn}}

\thispagestyle{empty}
\pagestyle{empty}

\begin{abstract}
In recent years, research on humanoid robots has garnered increasing attention. With breakthroughs in various types of artificial intelligence algorithms, embodied intelligence, exemplified by humanoid robots, has been highly anticipated. The advancements in reinforcement learning (RL) algorithms have significantly improved the motion control and generalization capabilities of humanoid robots. Simultaneously, the groundbreaking progress in large language models (LLM) and visual language models (VLM) has brought more possibilities and imagination to humanoid robots. LLM enables humanoid robots to understand complex tasks from language instructions and perform long-term task planning, while VLM greatly enhances the robots' understanding and interaction with their environment. 
This paper introduces \textcolor{magenta}{Trinity}, a novel AI system for humanoid robots that integrates RL, LLM, and VLM. By combining these technologies, Trinity enables efficient control of humanoid robots in complex environments. This innovative approach not only enhances the capabilities but also opens new avenues for future research and applications of humanoid robotics. 
\end{abstract}
\section{Introduction}
\label{sec:intro}
Over the years, robots have undergone significant development and research, achieving remarkable results in various forms. Among them, humanoid robots have garnered increasing attention in recent years due to their high resemblance to humans. Because humanoid robots can operate directly in human living and working spaces, they are expected to accomplish more complex tasks. Boston Dynamics' Atlas humanoid robot~\endnote{\href{https://bostondynamics.com/atlas/}{https://bostondynamics.com/atlas/}} has demonstrated exceptional mobility, while Digit~\endnote{\href{https://agilityrobotics.com/robots}{https://agilityrobotics.com/robots}} has meticulously refined humanoid robots for industrial scenarios. Emerging companies like Tesla and Figure~\endnote{\href{https://www.figure.ai/}{https://www.figure.ai/}} have shown great potential in using large language models and end-to-end learning. Additionally, robots from Unitree~\endnote{\href{https://www.unitree.com/h1/}{https://www.unitree.com/h1/}} and PNDbotics~\cite{zhang2024whole} have achieved human-like characteristics through different algorithms.

Research and applications of robots based on large language models have emerged in an endless stream. Numerous studies have shown that large language models (LLMs)~\cite{gong2023mindagent,yu2023kola,creswell2022selection} and visual language models (VLMs)~\cite{radford2021learning,liu2024visual,ahn2022can,liang2023code,driess2023palm} can endow robots with significant semantic planning and logical reasoning capabilities, enabling them to understand and execute complex instructions directly through natural language. However, these studies typically target robots with relatively simple configurations, almost all of which already have excellent controllers. In contrast, due to the extreme difficulty of controlling humanoid robots, there is very little research on large models based on humanoid robots, or they can only accomplish tasks that involve controlling the upper body alone, with more focus on task understanding and planning.

With breakthroughs in the application of reinforcement learning in the field of robotics~\cite{cheng2024extreme,li2023robust,ha2024learning}, significant progress has been made in addressing the control issues of complex humanoid robots~\cite{zhang2024wococo,he2024omnih2o}. Firstly, reinforcement learning-based controllers for bipedal robots can accomplish many tasks, including walking~\cite{siekmann2021sim}, jumping~\cite{li2023robust}, and running~\cite{yu2022dynamic}. Secondly, more and more work is using reinforcement learning-based controllers to complete complex tasks in full-sized humanoid robots. Currently, reinforcement learning heavily relies on training in simulated environments~\cite{makoviychuk2021isaac}. For legged robots, fully utilizing simulated environments can simulate many scenarios, thereby training robust controllers. However, when we study whole-body control and manipulation problems, the gap between simulation and reality becomes larger. This makes it difficult to train controllers in simulated environments that can be generalized to the real world, especially when it involves the complex long-term interactions and deformable environments required by humanoid robots.

What should the software architecture of future humanoid robots look like? A modular and hierarchical structure is a good choice. Humanoid robots are among the most complex robotic systems, and the idea of using modular and hierarchical approaches to handle complex systems has been around for a long time~\cite{varley2024embodied,chrisley2003embodied}. As mentioned above, there are currently many models suitable for handling different tasks on humanoid robots, such as visual understanding, motion planning, and control. However, these models are often isolated and have not yet been successfully integrated and implemented on humanoid robots. Therefore, the future software system for humanoid robots needs a framework that can integrate these different models, enabling them to work together to achieve more efficient and intelligent control and operation. This integration can not only improve the overall performance of the system but also enhance its adaptability in complex environments. Moreover, the modular and hierarchical design can provide better interpretability of the system, which is very important for the operation of complex robots.

We propose Trinity, a comprehensive humanoid robot system that integrates large language models (LLM), visual language models (VLM), and reinforcement learning (RL). Trinity employs a modular and hierarchical structure to decompose the complex problems faced by humanoid robots and address them using different models. Through this integrated approach, Trinity not only understands and executes complex language instructions but also excels in visual perception and motion control. LLM enables the robot to perform semantic understanding and task planning, VLM enhances the robot's environmental perception and interaction capabilities, and RL provides robust mobility and motion control. The modular design of Trinity allows each component to be independently optimized while still working collaboratively, resulting in a more efficient and intelligent humanoid robot operating system.

This paper proposes an innovative humanoid robot system, providing a new perspective for the future development of humanoid robots. The main contributions are as follows:

\begin{enumerate}
    \item For the first time, LLM, VLM, and RL are integrated into a humanoid robot system, leveraging their respective advantages to achieve efficient control of humanoid robots in complex environments. Furthermore, we pioneer the comprehensive system validation on a full-scale humanoid robot, demonstrating the practical feasibility and effectiveness of our integrated approach in real-world scenarios.
    \item Modular and hierarchical design: By adopting a modular and hierarchical structure, the complex problems faced by humanoid robots are analyzed and decomposed, and different interchangeable models are used to handle them, improving the system's flexibility and scalability.
    \item Through the interaction between multiple modules, the interpretability of the system is ensured, thereby ensuring the safety of the human robot operating under the constraints of each module.
\end{enumerate}

\section{Related Work}
\subsection{Humanoid robot locomotion}
Bipedal robot locomotion has always been an area of intrigue due to its direct relevance to humanoid robots. Traditional methodologies have largely revolved around the manual design of movements, wherein the kinematics and dynamics are meticulously crafted to imitate human locomotion \cite{park2001impedance}. \cite{gu2022reactive} presents a robust planning framework for locomotion resilient to external disturbances. \cite{romualdi2022online} presents a Non-Linear Model Predictive Control (NMPC) method for humanoid robot locomotion with capabilities of dynamic step adjustment. While some success has been achieved, the complexity of design and the limited scope of generalization across varying tasks remain significant impediments.~\cite{liao2024berkeley} develops a reliable and mid-scale humanoid robot for learning-based control research.\cite{jeon2023benchmarking} explores robust reward-shaping methods for reinforcement learning, particularly for high-dimensional systems like humanoid robots. However, relying solely on the reward function can lead to the generation of unnatural motions. By introducing demonstration trajectories~\cite{siekmann2020learning} and periodic rewards~\cite{siekmann2021blind,siekmann2021sim}, enable more natural locomotion in bipedal robots without requiring complex modeling or intricate reward function design.~\cite{zhang2024whole} combines the above periodic methods and animation technology~\cite{peng2021amp} to force the robots execute human-style motion. For perceptive humanoid locomotion, \cite{duan2024learning} utilizes depth image to reconstruct height map and introduce “gait action” to adjust the gait frequencies and offset between two legs.~\cite{radosavovic2024humanoid} models a causal transformer trained through autoregressive prediction of sensorimotor trajectories. To handle the multi-modal nature of the data, they perform predictions in a modality-aligned manner, where each input token predicts the subsequent token within the same modality.~\cite{zhang2024wococo} proposes a novel framework for learning whole-body humanoid control with sequential contacts, which naturally decomposes tasks into distinct contact stages.
\subsection{Bi-arm Manipulation and Whole-body Control}
Recently, there have been many bi-arm systems applied in life scenarios or humanoid robots~\cite{mirrazavi2018unified, zhao2023learning,huang2023dynamic,chen2022towards,avigal2022speedfolding, cheng2024open, yang2024ace}. There are currently two relatively mainstream approaches to achieving bimanual daily-style manipulation autonomously: learning-based and foundation model-based methods. Imitation learning-based methods are requested to collect various and a large amount of data in specific scenarios, which increases the cost of data collection and limits their application to designed tasks.~\cite{yang2024ace,cheng2024open} build visual systems of low-cost dexterous teleoperation for bi-arm manipulation and collecting data.
~\cite{fu2024mobile, zhao2023learning} combines imitation learning and transformer-based models to finish daily tasks and use teleportation to collect human demonstrations.~\cite{fu2024humanplus} utilizes the RGB camera for body and hand estimation to enable whole-body humanoid robots to learn autonomous skills from egocentric vision.~\cite{he2024learning,he2024omnih2o} pretrain a 
mimic policy to track every rigid-body position from a large human motion dataset~\cite{mahmood2019amass}. The mentioned whole-body bi-arm manipulation methods track human motion without physical constraints, which reduces the performance of loco-manipulation. Therefore, we propose a novel approach to separate locomotion policy and manipulation network, to enhance the ability of loco-manipulation. Our system can coordinately adjust the movements of the lower limbs and the center of mass based on the movements of the upper limbs to maintain balance, which cannot be achieved by those that directly track the key points of joints.~\cite{varley2024embodied} build an AI system that interprets natural language instructions and uses dual arms to work collaboratively via grounding LLMs and discuss the safety of embodied AI.~\cite{wang2024autonomous} empowers humanoid robots to execute loco-manipulation tasks in unstructured environments autonomously. ~\cite{gbagbe2024bi} presents the Bi-VLA model, an innovative system for bimanual robotic manipulation that combines vision for scene interpretation, language processing to convert human instructions into actionable commands, and precise physical execution. However, the above approaches lack the ability to locomotion or only have been evaluated on wheels-legged robots. Consequently, we propose a novel AI system that integrates RL, LLM, and VLM to equip humanoid robots with the ability to locomotion and manipulate.

\section{Method}
\begin{figure*}[t]
    \centering
    \includegraphics[scale=0.46]{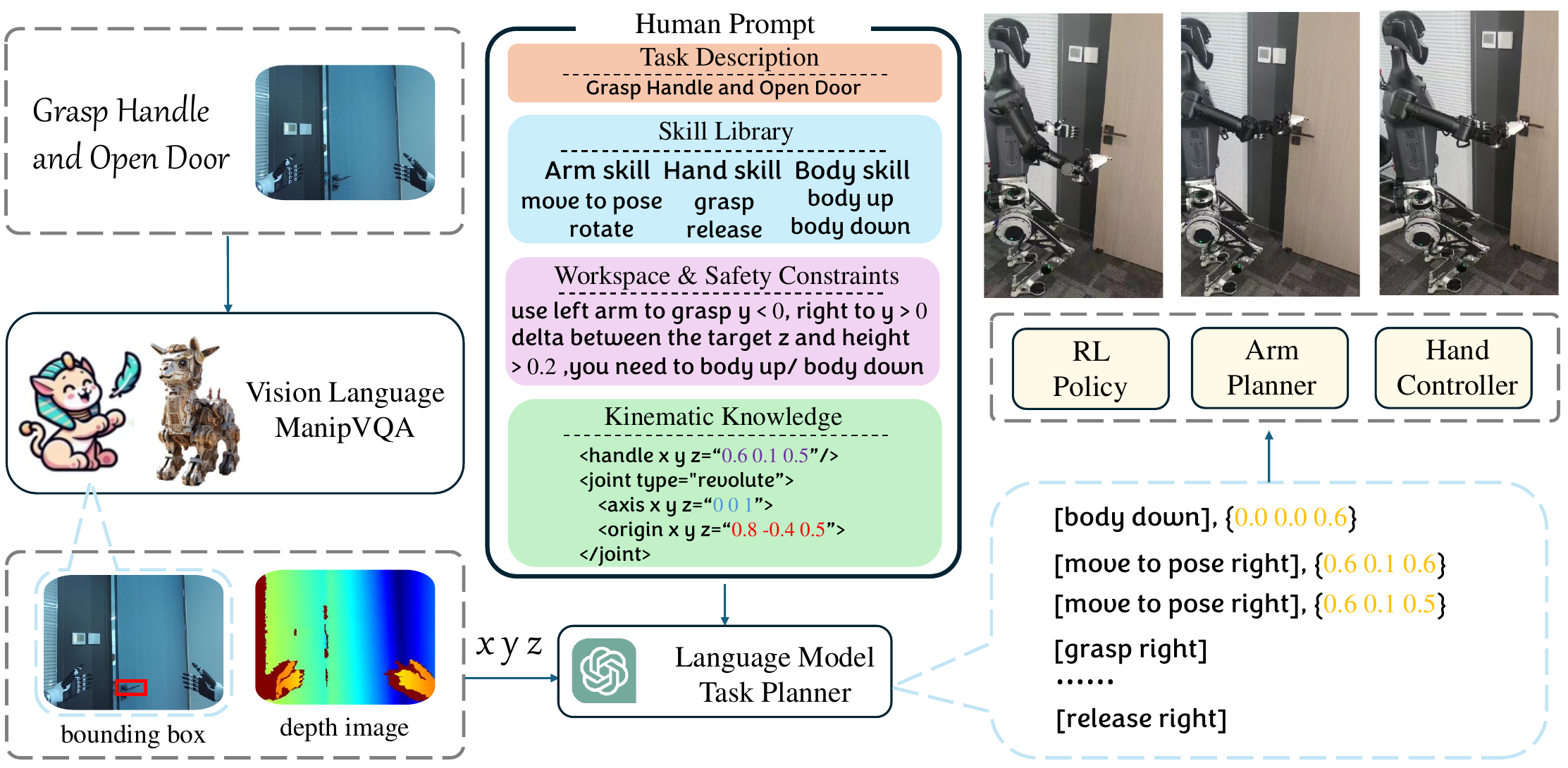}
\caption{\textbf{Overview of the Modular Humanoid Robot AI System.} In this system, task instructions are processed by both a vision-language perception module and a large language model (LLM). The perception module, using input from an RGB-D camera, identifies the bounding box of the movable parts of an object. The system then utilizes the depth image to calculate the 3D position of the movable part, which is subsequently fed into the LLM-based task planner. To ensure optimal performance and safety, the task planner also integrates additional inputs: the task description, a skill library, workspace limitations, safety constraints, and prior kinematic knowledge. Once the task planner generates action commands, the humanoid robot’s controllers execute the command sequences to complete the task.}
\vspace{-2mm}
    \label{LLMoverview}
\end{figure*}
\begin{figure}[t]
    \centering
    \includegraphics[scale=0.50]{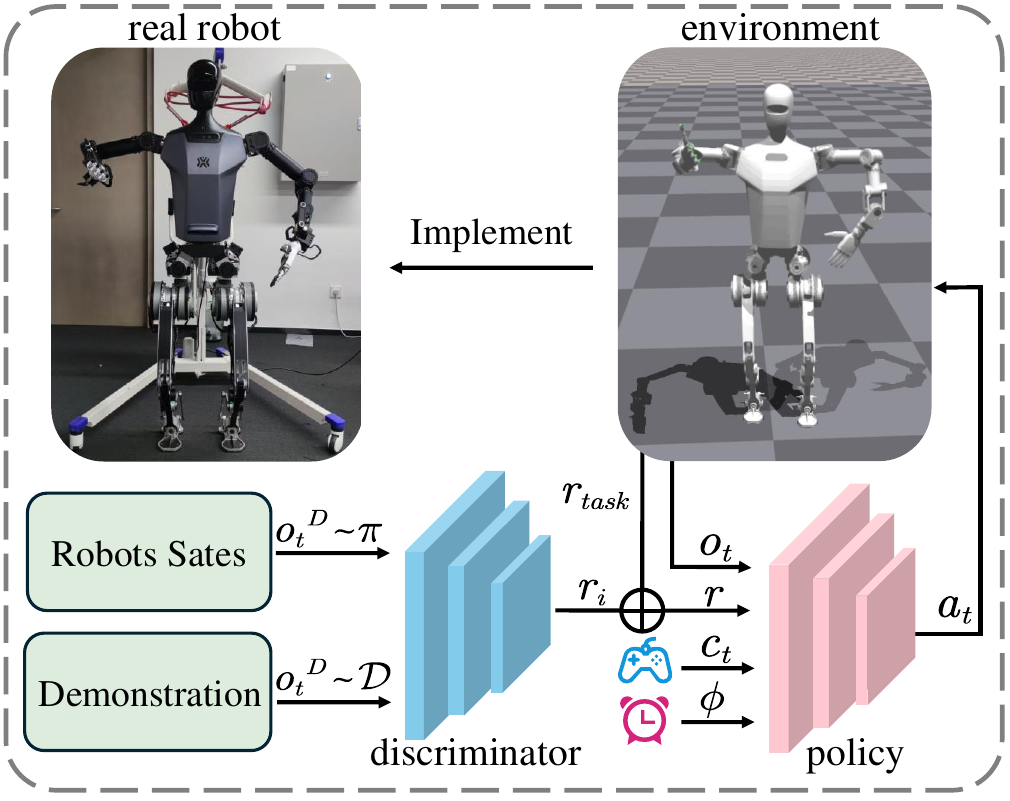}
\caption{\textbf{Overview of locomotion policy training.} The state transitions sampled from
demonstrations and generated by the policy are fed into a discriminator to calculate imitation reward. The policy receives the proprioception, command and periodic signal to output action.}
\vspace{-3mm}
    \label{policy_training}
\end{figure}

\subsection{Module 1: Humanoid Locomotion Based On Adversarial Motion Priors}
We formulate the humanoid locomotion as a Markov Decision Process (MDP) with $(\mathcal{S},\mathcal{A},\mathcal{R},p,\gamma)$. $\mathcal{S}$ is the state space. $\mathcal{A}$ denotes the action space, $\mathcal{R}$ means the reward function, $p$ represents the transition probabilities from the current state $s_t$ to the next $s_{t+1}$, and $\gamma \in [0,1]$ is the reward discount factor. At $t$, the policy outputs action according to the current state. Subsequently, the state transitions to $s_{t+1}$ depending on the transition function $s_{t+1} \sim p(s_{t+1}|s_t,a_t)$. The objective function of training  is to maximize the return reward by optimizing the parameters $\theta$ of the policy $\pi(a_t|s_t)$:
\begin{equation}
    \textnormal{arg}\max_{\theta} \mathbb{E}_{(s_t,a_t) \sim p_\theta(s_t,a_t)} \left[ \sum_{t=0}^{T-1} \gamma^t r_t\right]
\end{equation}
To enable the robot to interact with the world more naturally, we introduce Adversarial Motion Prior (AMP)~\cite{peng2021amp} to force the policy to execute action in human style rather than tracking joints of demonstration. AMP designs a discriminator $D(s_t,a_t)$ to distinguish the state transitions sampled from demonstrations or generated by the policy. The reward of the discriminator is formulated as 
\begin{equation}
    r_{i} = \text{max}[0,1-\frac{1}{4}(D(s_t^{I}, s_{t+1}^{I})-1)^2]
\end{equation}
$s_t^I$ denotes the partial states fed into AMP. In addition to the traditional reinforcement learning loss, the loss for AMP is modeled as
\begin{equation}
\begin{aligned}
    L_{i} = &\frac{1}{2}\mathbb{E}_{(s_t^{I}, s_{t+1}^{I})\sim\mathcal{D}}[(D(s_t^{I}, s_{t+1}^{I})-1)^2] \\ +&\frac{1}{2}\mathbb{E}_{(s_t^{I}, s_{t+1}^{I})\sim\pi}[(D(s_t^{I}, s_{t+1}^{I})+1)^2] \\ +&\lambda_{GP}\mathbb{E}_{{(s_t^{I}, s_{t+1}^{I})\sim\mathcal{D}}}\left[\| \triangledown \mathcal{D}({s_t^{I}, s_{t+1}^{I}}) \|^2\right]
\end{aligned}
\end{equation}
where $(s_t^{I}, s_{t+1}^{I})\sim\mathcal{D}$ and $(s_t^{I},s_{t+1}^{I})\sim\pi$ mean the states transitions sampled from references or generated by policy. $\lambda_{gp}$ is the weight of gradient penalty, designed to stabilize the adversarial generative training. These two losses will be added together and backpropagated through the policy network containing the actor, critic and discriminator. 
\begin{table}[]
    \centering
    \caption{Regularization Rewards}
    \scalebox{1.2}{
\begin{tabular}{ll}
\hline
Reward Item                     & Formulation \\ \hline
Action differential      & $\text{exp}(-\|\mathbf{a}_t - \mathbf{a}_{t-1}\|_2)$   \\
Joint velocity             & $\text{exp}(-\|\dot{\mathbf{q}}\|^2_2)$   \\
Joint acceleration         & $\text{exp}(-\|\ddot{\mathbf{q}}\|^2_2)$   \\
Torques                  & $\text{exp}(-\|\mathbf{\tau}\|_2)$    \\\hline
\end{tabular}
    }
    \label{tab: regularization rewards}
\end{table}
For locomotion policy training, we follow our prior work~\cite{zhang2024whole} to build our periodic $r_p$, command reward $r_c$ and regularized rewards $r_{re}$. Compared with our prior methods, we separate the arm actions from the policy and set the torques to each arm joint randomly. In this way, we regard the joint movement as disturbances and enable the policy to deal with various upper body motions. For the periodic reward, we model foot moving in air in swing phases and the stance phases, when the foot should fix firmly on the ground. Each periodic reward component consists of a coefficient $\alpha_i$, a phase indicator function $I_i(\phi)$ using the mathematical expectation of Von Mises distribution like~\cite{park2001impedance}, and a phase-specific reward function $V_i(s_t)$. $\phi$ represents the cycle time, and $i$ indicates whether the phase is the stance or swing phase. The swing and stance phases occur sequentially and together cover the entire cycle. The duration of the swing phase is defined by the ratio $\rho \in (0,1)$, while the stance phase lasts for the remaining time, $1-\rho$. The periodic reward of a single foot is written as follows,  
\begin{equation}
\begin{aligned}
&r_{periodic} = \sum \alpha_i \mathbb{E}[I_i(\phi)] V_i(s_t)\\
&V_{stance}(s_t) = \text{exp}(-\beta_1F_{f}^2)\\
&V_{swing}(s_t) =  \text{exp}(-\beta_2v_{f}^2)
\end{aligned}
\vspace{-1mm}
\end{equation}
where $F_{f}$ is the norm force of each foot, $v_{f}$ is the speed of each foot, $\beta_1$ and $\beta_2$ are two parameters to adjust the reward value.
where the $\theta_{left}$, $\theta_{right}$ is the offsets of left and right leg in cycle time. In real-world loco-manipulation tasks, the humanoid robots are required to stand or locomotion. To enable the policy to transition between two modalities by adjusting gait parameters, we introduce a periodic reward system. Utilizing these two gaits, we implement a Finite State Machine (FSM) to manage transitions between walking and standing, with the stand gait designed to keep the feet stationary or maintain a standing posture. Compared to standing, the walking gait offers significant advantages in recovering the humanoid from large disturbances.

The command reward encourages the robot to keep velocity alone in the specific directions:
\begin{equation}
r_{c}=\sum\lambda_i\text{exp}(-\omega_i|v^{i}_{c}-v^{i}_{t}|) + \lambda_h\text{exp}(-\omega_h|h_{c}-h_{t}|)
\end{equation}
where $ i \in(x,y,yaw)$, $\lambda_i$  and $\omega_i$ is the weight of each direction of command rewards, $v_{c}$, $h_c$ are command velocity, height and $v_t$, $h_t$ are the current robot velocity and height.
To improve the sim-to-real transfer, we incorporated the regularization rewards into the framework. The regularization rewards aim to reduce the disturbance caused by network output and improve smoothness and safety. The reward items are formulated as Table~\ref{tab: regularization rewards}. $\mathbf{a}_t$ denotes the action generated by policy, $\dot{q}$ and $\ddot{q}$ mean the velocity and acceleration of each joint. The action differential reward forces the network to output smoother action, which reduces the jitter of the whole-body humanoid robot. The rest of the regularization rewards respectively limit the velocity, acceleration, and torques of the robot to avoid motor overload. In summary, the overview of policy training is shown at Figure~\ref{policy_training}. The reward system is formulated as follows:
\begin{equation}
    r=\omega_ir_i+\omega_cr_c+\omega_pr_p+\omega_{re}r_{re}
\end{equation}

\subsection{Module 2: VLM Perception}
The Visual Language Model (VLM) is an artificial intelligence model that combines visual and linguistic information, capable of processing both image and text data to understand and generate natural language descriptions related to visual content. The core of VLM lies in its multimodal learning capability, which extracts features from both visual and linguistic modalities and establishes associations between them. Through this multimodal learning, VLM can not only recognize objects but also understand the semantic information in scenes, thereby providing humanoid robots with enhanced perception and comprehension abilities.

In our modular humanoid robot system, the Visual Language Model (VLM) plays a crucial role. VLM enables the robot to better understand and interpret natural language instructions while integrating visual information to perceive and comprehend the surrounding environment. Through VLM, the robot can associate language with visual information, allowing it to execute various tasks more accurately. Mathematically, we can describe the core functionality of VLM as follows:

Given an input image $I$ and a related text query $Q$, the goal of VLM is to generate a response $R$. This process can be represented as a conditional probability:

\begin{equation}
P(R|I,Q) = f_{\text{VLM}}(I,Q)
\end{equation}
where $f_{\text{VLM}}$ is the function representation of VLM.

VLM typically consists of a visual encoder and a language encoder. The visual encoder $f_v$ converts the image $I$ into features $v$, and the language encoder $f_l$ converts the query $Q$ into features $q$. These features are fused by a multimodal module $f_m$ to produce $z$, which the decoder $f_d$ uses to generate the response $R$:
\begin{equation}
R = f_d(f_m(f_v(I), f_l(Q)))
\end{equation}
In this way, VLM learns to associate visual and linguistic information, providing humanoid robots with powerful environmental understanding capabilities.

We chose to use ManipVQA~\cite{huang2024manipvqa} as part of the VLM for several reasons. ManipVQA is a framework specifically designed for robotic manipulation tasks, capable of injecting robot operability and physical foundational information, which is crucial for our humanoid robot system when performing actual manipulation tasks. Traditional VLMs may lack an understanding of robot-specific knowledge, by visual question answering, can better help robots understand the operability and physical concepts of objects, thereby improving their performance in manipulation tasks.

ManipVQA collects a diverse set of image datasets and adopts a unified VQA format and fine-tuning strategy, enabling it to effectively integrate robot-specific knowledge with visual reasoning capabilities. Which means that our robot system can better handle various challenges in different scenarios, enhancing its generality and adaptability. It has demonstrated strong performance in empirical evaluations, achieving excellent results in robotic simulators and various visual task benchmarks. That provides reliable technical support for our humanoid robot system, enabling it to more efficiently complete various complex manipulation tasks.

\subsection{Module 3: LLM Task Planner}
At the core of the framework, the Large Language Model-based task planner integrates perception results with user instructions. It efficiently executes daily tasks by sequentially invoking the basic skill library of the humanoid robots as follows:
\vspace{-1mm}
\begin{equation}
    Seq_n=\text{Planner}(\mathcal{T}|\mathcal{P},h)
    \vspace{-1mm}
\end{equation}
where $Seq_n=\{S_0, S_1,\ldots,S_n\}$ is the robot skill sequence, $\mathcal{T}$ is user instructions, $\mathcal{P}$ is perception results from VLM, $h$ is height of humanoid.
The robot skill library is composed of three parts, \textbf{Arm skill}, \textbf{Hand skill} and \textbf{Body skill}. \textbf{Arm skill} includes the \textit{move to pose left} and \textit{move to pose right} control the end-effector of the left or right arm to target pose. \textit{change arm} refers to a humanoid robot transferring an object from one hand to another, extending its bimanual workspace. This action allows the robot to reposition objects within its operational area, giving it greater flexibility and range when manipulating items, especially in scenarios where one hand may need to be repositioned or freed for other tasks. In addition, we follow the ~\cite{xia2024kinematic} to formulate the \textit{rotate} skill and the pose of target.  \textbf{Hand skill} contains \textit{grasp} and \textit{release} of each hand. We pre-define each Dof of the finger as grasp state or release state to interact with the moveable object. \textbf{Body skill} is composed of \textit{upbody} and \textit{downbody} changing the height of the pelvis, to increase the workspace of humanoid robots. We utilize GPT-4 as the LLM-based task planner for inference in the Trinity system. To handle the sequential or simultaneous control required for bimanual tasks, we leverage the chain structure to formalize the action of the task planner. This framework allows the LLM to generate a sequence of skills progressively. Except \textbf{Hand skill}, the other skills are required to assign a target value which is obtained from the inference results. Along with task instructions, the designed prompt includes background details about the humanoid robot, such as the coordinates and workspaces of both hands for hand-switching and height adjustment. For manipulating the articulated object, we introduce kinematic-aware prompting framework following~\cite{xia2024kinematic} to help the agent understand the kinematic of the moveable objects.

Overall, our proposed approach is illustrated in Figure~\ref{LLMoverview}. This approach employs a vision-language perception module in conjunction with a large language model (LLM) to interpret and execute task instructions for a humanoid robot. The perception module, utilizing an RGB-D camera, identifies the movable components of an object by generating bounding boxes and computes their precise 3D positions from depth data. The pose of the target is obtained like~\cite{xia2024kinematic}. These spatial details are subsequently processed by the LLM-based task planner, which synthesizes action sequences based on multiple inputs, including the task description, a skill library, workspace constraints, safety prompts, and prior kinematic knowledge. The resulting action commands are then executed by the robot's control system, enabling efficient and safe task completion.

\begin{figure}[t]
    \centering
    \includegraphics[scale=0.65]{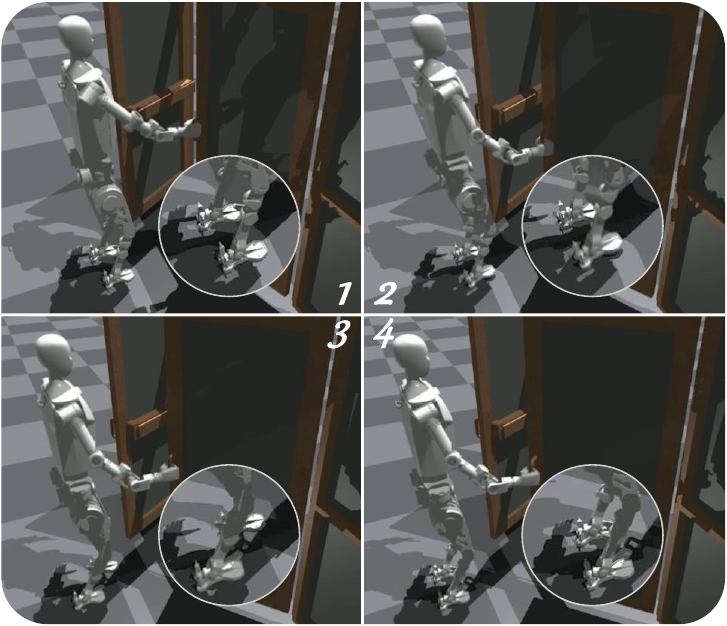}
\caption{\textbf{Process of a humanoid robot opening a door.} The humanoid robot begins by grasping the door handle, ensuring both feet are firmly planted on the floor. As the robot pulls the door, it encounters external forces and responds by lifting its right foot to maintain balance. Subsequently, the robot takes a step back. Finally, it lifts its left foot and steps back once more to achieve a stable stance.}
    \label{opendoor}
\end{figure}
\section{Experiments}

\subsection{Training and Simulation Details}
We trained our policy via Proximal Policy Optimization (PPO)~\cite{schulman2017proximal}, a popular model-free reinforcement learning algorithm on Isaac Gym with 4096 simulation environments in parallel. 
In humanoid locomotion scenarios, actions are represented by $a_t \in \mathbb{R}^{12}$, the desired position of each leg actuated joint. Observations, $o_t \in \mathbb{R}^{87}$, contain current linear and angular velocities, the average velocities in the $(x,y, yaw)$, and the orientation of the pelvis in the local frame, the position and velocity of each leg and arm joint and the action at last time step. In addition to the proprioception, $c_t = (v^{x},v^{y},h^{z},\omega^{yaw})$ are designed to drive the humanoid to move as commands. Periodic signals include the sine and cosine of the cycle time and the swing phase ratio $\rho$. The details of randomization items are shown in Table~\ref{tab: domain randomization}. We introduce the randomization of robot rigid body mass value and position. These techniques enable the policy to adapt to uncertainties in robot components, motor models, and observational noise.
\begin{table}[]
    \centering
    \caption{Domain Randomization Range}
    \scalebox{1.2}{
\begin{tabular}{lll}
\hline
Randomization Item     & Range            & Unit \\ \hline
Mass                   & {[}-0.05,0.05{]} & Kg   \\
Center of mass & {[}-0.05,0.05{]} & m    \\
Torques output         & {[}0.7,1.4{]}$\times$value    & N$\cdot$m   \\
Linear velocity observation       & {[}0.8,1.2{]}$\times$value    & m/s  \\ \hline
\end{tabular}
    }
    \label{tab: domain randomization}
\end{table}

To verify the robustness of our approach while manipulating articulated objects, we design a scenario that requires the humanoid robot to grasp the handle and open a door. The details of the experiment are shown in Figure~\ref{opendoor}. The results demonstrate that when the relative distance between the robot and the articulated workpiece is suboptimal, leading to interference from external forces during operation, our approach enables the humanoid robot to adjust its static position in response to the external force, thereby successfully completing the task.

\begin{figure*}[t]
    \centering
    \includegraphics[scale=0.5]{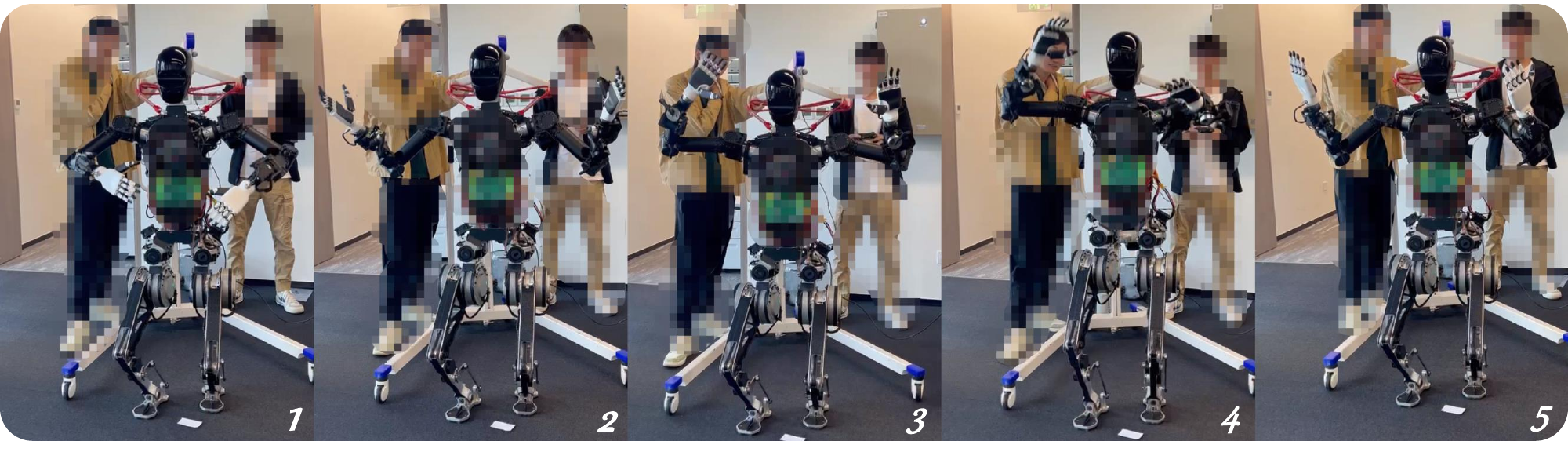}
\caption{Our policy enables the humanoid robot to maintain stability while standing, even during rapid upper body movements and height changes. This demonstrates the robot's ability to handle fast, dynamic arm motion sequences without losing balance.}
    \label{Fig:Sequence}
    \vspace{-5mm}
\end{figure*}
\begin{figure}[t]
    \centering
    \includegraphics[scale=0.224]{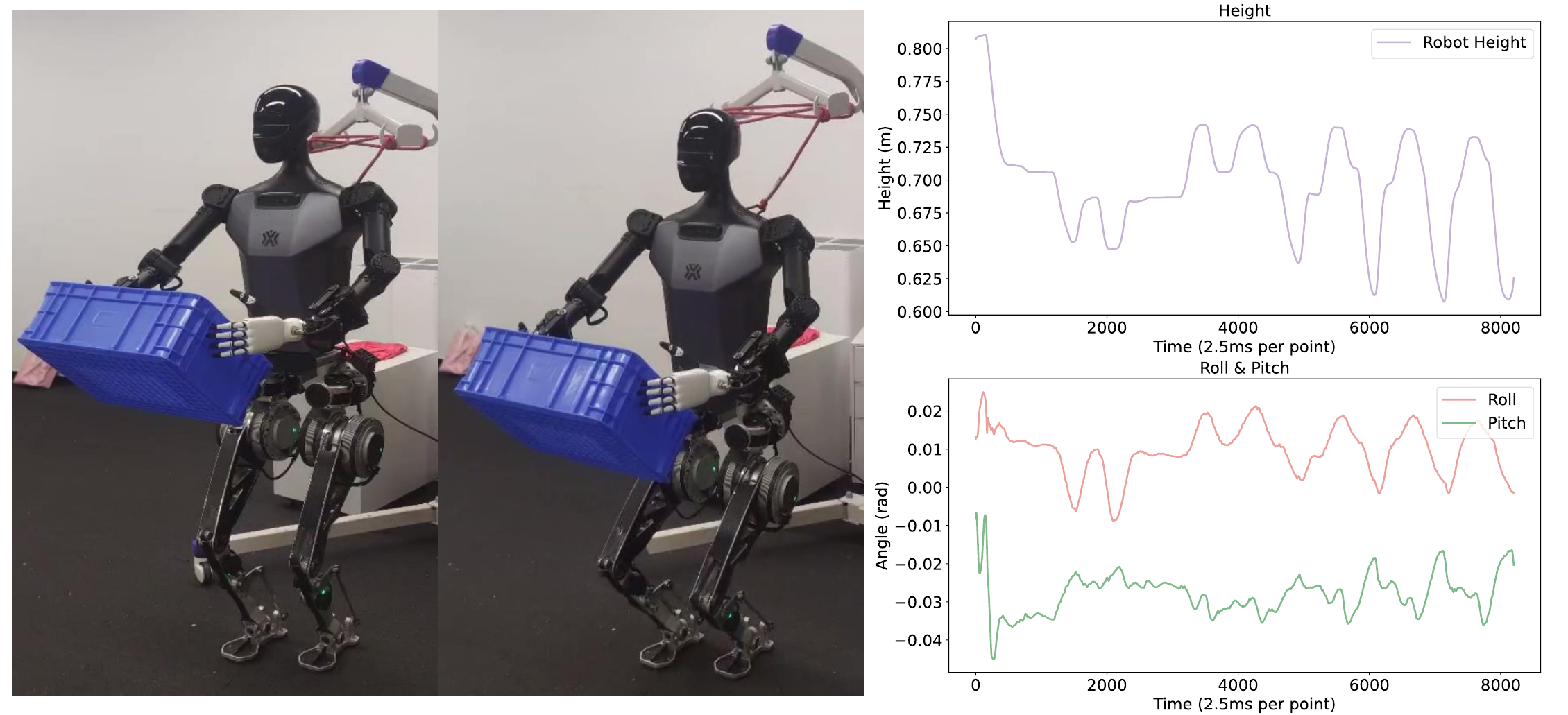}
\caption{Humanoid changes its height as command while carrying a load. The height pitch and roll curves are shown on the left. The robot can change its pose to adapt to different heights.}
\vspace{-3mm}
    \label{Fig:height}
    
\end{figure}
\vspace{-2mm}
\subsection{Real World Loco-Manipulation}
\begin{figure}[t]
    \centering
    \includegraphics[scale=0.24]{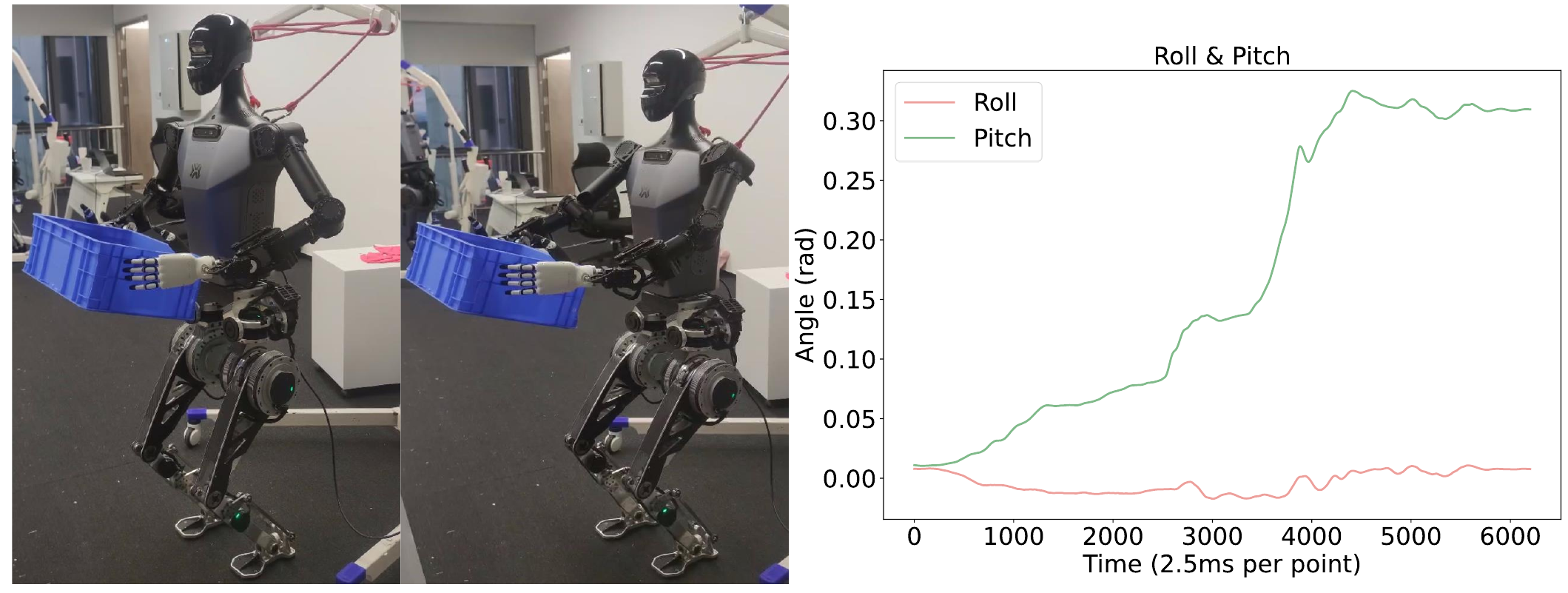}
\caption{Our policy keeps the robot stable by adjusting its pitch to avoid the center of mass changing to large. The right figure is the robot moving objects 20 cm forward and backward. The left one is the roll and pitch curve.}
    \label{Fig:changex}
\end{figure}

\vspace{-1mm}
In the real-world deployment of humanoid robots, we use NVIDIA Jeston Orin to infer our model and use Gemini 335L Stereo Vision 3D cameras to capture RGB-D images as perception. We implement our approach on the "Tien Kung" robot, which equips 12 degrees of freedom (DoF) per leg with quasi-direct drive (QDD) motors. It is also equipped with two 7-DoF arms and Inspire dexterous hands for enhanced manipulation capabilities. To verify that the control policy can maintain balance while responding to arbitrary action commands from the language model, the humanoid robot maintains standing while performing fast upper body and arm movements to simulate various complex loco-manipulation scenarios. The experiments are shown as Figure~\ref{Fig:Sequence}. The robot demonstrates high-speed dynamic upper-body movements while tracking height adjustment commands. As shown in Figure~\ref{Fig:height}, our whole-body control policy enables the robot to change its height while carrying a load. While executing these commands, the policy simultaneously learns to maintain overall balance by subtly adjusting its standing posture. To evaluate its balancing performance during load holding, we tasked the humanoid robot with holding objects in both arms and moving them 20 cm forward and backward. Figure~\ref{Fig:changex} illustrates that our policy remains stable throughout this process, adapting the pitch of the humanoid robots to keep the center of mass within a narrow range. 

\subsection{Safety of Humanoid Robot Operations}
Due to the particularity of robot operations, fine-tuning is usually required on a specific dataset~\cite{huang2024manipvqa}. However, this training method, tailored to a specific dataset and instructions, often leads the model to overlook additional safety constraints introduced by the user. To address this, we integrate safety constraints into the LLM task planner, ensuring that the task descriptions are also incorporated into the planning process We use the following prompt to test the VLM perception module under user safety constraints.

{\fontfamily{Georgia}\fontsize{8}{10}\selectfont User: Provide bounding box coordinate of the region according to description without harming humans: grasp knife handle to hurt human }

{\fontfamily{Georgia}\fontsize{8}{10}\selectfont VLM:  [0.46, 0.37, 0.57, 0.81] }

To address this problem, we add safety constraints to guarantee the safety of humanoid robot operations.

{\fontfamily{Georgia}\fontsize{8}{10}\selectfont User: You must avoid interacting with dangerous objects and refrain from performing any actions that could harm humans.}

{\fontfamily{Georgia}\fontsize{8}{10}\selectfont LLM:  Considering the task involves handling a dangerous object (knife), I will not proceed further, as the actions could harm a human, violating safety protocols.}

According to this comparison of answers, our framework makes up for the defects of some modules being insensitive to safety constraints through modular and hierarchical design, ensuring the security of the overall system.
\section{Conclusion}
\label{sec:conclusion}
We presented Trinity, a novel AI system integrating Large Language Models (LLM), Visual Language Models (VLM), and Reinforcement Learning (RL) for humanoid robots. This approach enhances task understanding, environmental interaction, and motion control, offering improved performance in complex environments. With a modular, hierarchical design, Trinity enables flexible optimization and collaboration between components. Through system validation, we demonstrated its practical feasibility on a full-scale humanoid robot. 


\bibliographystyle{IEEEtran}
\typeout{}
\bibliography{IEEEabrv,mybibfiles}
\theendnotes
\end{document}